\documentclass{article}

\PassOptionsToPackage{numbers, compress}{natbib}

\usepackage[final]{neurips_2025}

\usepackage[utf8]{inputenc} % allow utf-8 input
\usepackage[T1]{fontenc}    % use 8-bit T1 fonts
\usepackage{hyperref}       % hyperlinks
\usepackage{url}            % simple URL typesetting
\usepackage{booktabs}       % professional-quality tables
\usepackage{amsfonts}       % blackboard math symbols
\usepackage{nicefrac}       % compact symbols for 1/2, etc.
\usepackage{microtype}      % microtypography
\usepackage{xcolor}         % colors
\usepackage{graphicx}
\usepackage{booktabs}
\usepackage{multirow}
\usepackage{amsmath} 
\usepackage{booktabs} % For formal tables
\usepackage{multirow}
\usepackage{pifont}
\usepackage{xcolor}
\usepackage{graphicx}
\usepackage{subfig}     % For subfloat
\usepackage{amsmath}

\usepackage{array}

\usepackage{mathtools}
\usepackage{algpseudocode}
\usepackage{listings}
\usepackage{wrapfig}
\usepackage{colortbl}
\usepackage{amssymb}
\usepackage{bbm}
\usepackage{gensymb}
\usepackage{amsfonts}
\usepackage{colortbl} 
\usepackage{caption}

\usepackage{overpic}
\usepackage{setspace}
\usepackage{textpos}
\usepackage{enumitem}

\definecolor{citecolor}{HTML}{2980b9}
\definecolor{linkcolor}{HTML}{c0392b}
\usepackage{hyperref}
\hypersetup{hidelinks,breaklinks=true,colorlinks,citecolor=citecolor,linkcolor=linkcolor}

\newcommand{\tabincell}[2]{\begin{tabular}{@{}#1@{}}#2\end{tabular}}
\usepackage{wrapfig}
\title{VTBench: Comprehensive Benchmark Suite Towards Real-World Virtual Try-on Models}

\author{
  \textbf{Xiaobin Hu}$^{1}$,
  \textbf{Yujie Liang}$^{1,3}$,
  \textbf{Donghao Luo}$^{1,2}$,
  \textbf{Xu Peng}$^{1}$,
  \textbf{Jiangning Zhang}$^{1}$, \\
  \textbf{Junwei Zhu}$^{1}$,
  \textbf{Chengjie Wang}$^{1}$,
  \textbf{Yanwei Fu}$^{2}$ \\
  $^1$Tencent\quad
  $^2$Fudan University  \quad 
  $^3$Xiamen University \quad   \\
  \vspace{-9mm}
}
\begin{document}

\maketitle

\centerline{\qquad \textbf{\color{magenta} Project Website}: \url{https://github.com/HUuxiaobin/VTBench}} 

\begin{figure}[htbp]
\centering
\includegraphics[width=1\textwidth]{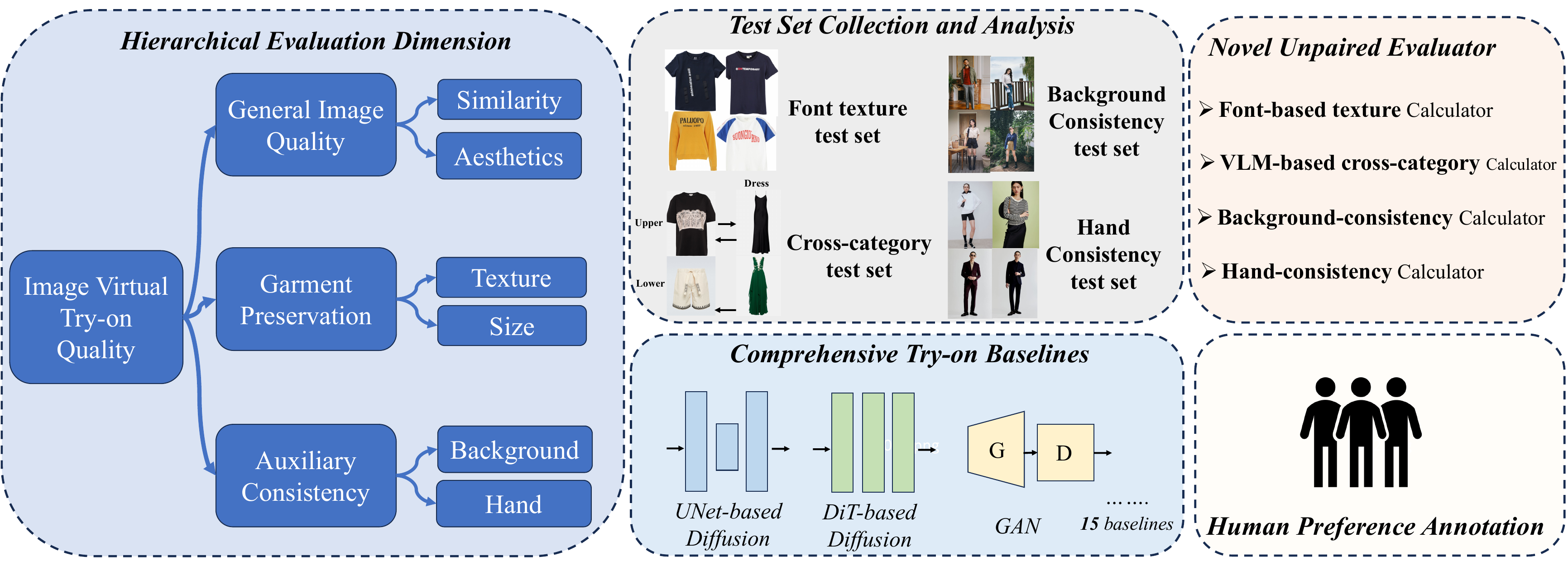}
% \vspace{-7mm}
\caption{\textbf{Overview of VTBench}. We propose VTBench, the first comprehensive benchmark suite designed to evaluate image-based virtual try-on models. 
To enable fine-grained and objective assessment, we propose a comprehensive and hierarchical Evaluation Dimension Suite that systematically  decomposes “image virtual try-on quality” into multiple well-defined dimensions. 
For each dimension and content category, we  curate a dedicated test set and develop reliable metrics, and then sample virtual try-on images from 15 virtual try-on models based on different foundations to provide in-depth insights.  We also conduct human preference annotation for  virtual try-on results across all dimensions, demonstrating strong alignment between VTBench’s automated evaluations and human perceptual judgments. Our benchmark delivers multi-perspective insights, advancing the systematic assessment of virtual try-on technologies.
}
\label{fig:teaser}
% \vspace{-3mm}
\end{figure}

\begin{abstract}
While virtual try-on has achieved significant progress, evaluating these models towards real-world scenarios remains a challenge. A comprehensive benchmark is essential for three key reasons: (1) Current metrics inadequately reflect human perception, particularly in unpaired try-on settings; (2) Most existing test sets are limited to indoor scenarios, lacking complexity for real-world evaluation; and (3) An ideal system should guide future advancements in virtual try-on generation.
To address these needs, we introduce VTBench, a hierarchical benchmark suite that systematically decomposes virtual image try-on into hierarchical, disentangled dimensions, each equipped with tailored test sets and evaluation criteria. VTBench exhibits three key advantages: 1) Multi-Dimensional Evaluation Framework: The benchmark encompasses five critical dimensions for virtual try-on generation (\textit{e.g.,} overall image quality, texture preservation, complex background consistency, cross-category size adaptability, and hand-occlusion handling). Granular evaluation metrics of corresponding test sets pinpoint model capabilities and limitations across diverse, challenging scenarios. 2) Human Alignment: Human preference annotations are provided for each test set, ensuring the benchmark’s alignment with perceptual quality across all evaluation dimensions. 3) Valuable Insights: Beyond standard indoor settings, we analyze model performance variations across dimensions and investigate the disparity between indoor and real-world try-on scenarios. To foster the field of virtual try-on towards challenging real-world scenario, VTBench will be open-sourced, including all test sets, evaluation protocols, generated results, and human annotations.
\end{abstract}

\section{Introduction}
\label{sec:intro}
Image-based virtual try-on (VTON) \cite{dong2019towards, ge2021disentangled, issenhuth2020not, han2019clothflow, han2018viton, he2022style, kim2024stableviton, minar2020cp, wang2018toward, xie2023gp, yang2020towards,yang2024texture, zhu2024m} is a widely adopted and highly promising image synthesis technology in the e-commerce industry. Its primary goal is to enhance the shopping experience for consumers and minimize advertising expenses for clothing merchants. The VTON task involves generating an image of a human model wearing a specific garment. Over the past few years, countless researchers 
%wang2018toward ## #
have dedicated significant efforts, such as Generative Adversarial Networks \cite{goodfellow2020generative} and Diffusion models \cite{rombach2022high}, to achieve more realistic and precise virtual try-on results.
As image virtual try-on models advance, there is an urgent demand for robust evaluation. These evaluation methods must not only align with human perceptual judgments of generated try-on results but also provide reliable metrics for assessing model performance. Furthermore, they should reveal individual models' specific capabilities and limitations, offering actionable insights to guide future improvements in data selection, training strategies, and model architecture selection, which ultimately drives progress toward more sophisticated real-world applications.

Existing virtual try-on evaluations usually adopt the paired evaluation metrics and replace the garment-agnostic model with the target garment. But it fails to timely assess the performance of virtual try-on in unconstrained settings without the guidance of paired ground-truths. 
Moreover, acquiring paired data is inherently challenging, especially when attempting to test arbitrary clothing items on diverse models across varied real-world scenarios. The complexity makes such data collection exceptionally difficult. Consequently, there exist strong desires of unpaired evaluators that are universal and effective to reliably reflect the quality of virtual try-on in open and challenging real-world scenarios getting rid of the strict paired shooting conditions.  

Currently, existing metrics for image virtual try-on generation mainly consists of paired metrics, such as Structural Similarity (SSIM) \cite{wang2004image} and Learned Perceptual Image Patch Similarity (LPIPS) \cite{zhang2018unreasonable}, and unpaired metrics, for example, Frechet Inception Distance (FID) \cite{parmar2022aliased} and Kernel Inception Distance (KID) \cite{binkowski2018demystifying}. 
These unpaired metrics are inconsistent with human evaluations and also neglect the unique, challenging characteristics of the virtual try-on field in real-world scenes. 
Hence, there is a pressing need for an evaluation framework that aligns closely with human perception and is specifically designed for the characteristics of virtual try-on models. To this end,  we introduce VTBench, a comprehensive benchmark suite for evaluating virtual try-on model performance. VTBench has three appealing properties: 1) comprehensive evaluation dimensions and corresponding test sets catering for difficult try-on challenging cases, 2) Novel unpaired reliable evaluators highly close with human alignment, and 3) building the comprehensive and most recent SOTA baselines, and providing valuable insights. 

First, our pipeline incorporates a hierarchical evaluation framework that systematically decomposes virtual try-on image quality through a structured and disentangled approach. The framework organizes the assessment into three fundamental dimensions: General Image Quality, Garment Preservation, and Auxiliary Consistency, each of which is subsequently refined into more specific evaluation criteria. This multi-level architecture ensures precise isolation and independent assessment of each quality aspect, eliminating potential interference between evaluation variables, as demonstrated in Fig. \ref{fig:teaser}.
For example, in terms of \textit{Garment Preservation}, we disentangle this coarse dimension into the finer ones including the texture and size to evaluate try-on model's capabilities in texture maintenance and cross-categories cases (\textit{e.g.,} dress$\leftrightarrow$any). 
Each finer evaluator is supported by high-quality self-collected labor-consuming dataset. 
Specifically, to evaluate the ability of texture maintenance, we geniously convert texture judgments into font structure texture evaluations and collect the garments with different size of font. 
We carefully collect  four test sets customized for corresponding four finer dimensions in the aspects of texture and size garment, and the auxiliary background and hand-occlusion consistency. 

% \leftrightarrow ///
Secondly, each granular dimension is supported by novel unpaired specifically-customized metrics overcoming the considerable difficulty of paired try-on dataset collections. A total of four novel unpaired evaluators are formulated to comprehensively verify the model's capabilities in six granular dimension, breaking the great barriers of current try-on benchmark only focusing on the similarity and distribution gap. 
Moreover, we systematically demonstrate that our evaluators exhibit strong alignment with human perception across all fine-grained evaluation dimensions. Specifically, we use the image try-on models to sample synthetic try-on results, and then arrange human annotators to give users' preferences based on the six granular dimensions. 
We demonstrate that VTBench evaluations exhibit a strong correlation with human preferences.

Lastly, VTBench's multi-dimensional approach compared with previous single and limited try-on benchmarks can provide valuable insights to the virtual try-on community towards more complicated real-world scenarios. For example, 
through our hierarchical dimension system, we provide granular feedback on the capabilities and limitations of virtual try-on across diverse evaluation dimensions. 
This framework not only thoroughly assesses existing models but also illuminates the path toward training sophisticated try-on generation systems, offering guidance to refine architectures and data strategies for higher-quality outputs.

We hereby open-source VTBench in its entirety, comprising a comprehensive suite of evaluation dimensions, novel evaluation method, curated test collections, generated virtual try-on results, and annotated human preference datasets. The research community is cordially invited to benchmark their image-based virtual try-on generation models against this standardized evaluation framework.

\section{Related Works}
\label{sec:formatting}
\noindent \textbf{Image Virtual Try-on Models.}
Over the past decade, image-based virtual try-on \cite{cui2024street} has garnered significant research attention, presenting itself as an increasingly promising field of remarkable e-commerce, which enhances the shopping experience for consumers.
A series of studies based on Generative Adversarial Networks (GANs)\cite{ge2021parser,issenhuth2020not,lee2022high,li2021toward,xie2023gp,yang2023occlumix,chopra2021zflow,chen2023size}, UNet-based Diffusion \cite{chen2024wear, liang2024vton, kolors,zhu2023tryondiffusion,choi2024improving,li2024anyfit}, and DiT-based Diffusion structures \cite{luo2025crossvton, jiang2024fitdit,ni2025itvton,luan2025mc,wan2024ted,zheng2024viton} have emerged for high-fidelity and high-realism image virtual try-on.  
The exponential growth of virtual try-on generation models underscores the urgent need for standardized evaluation protocols that can both analyze existing capabilities and inform future research directions. In this context, VTBench emerges as a pioneering solution, providing a meticulously designed benchmark suite that facilitates hierarchical, human-aligned performance analysis towards more challenging real-world try-on scenarios.

\noindent \textbf{Evaluator of Virtual Try-on Generative Models.} 
Existing virtual try-on generation models typically use metrics as follows:  In paired try-on settings with ground truth in test datasets, the reconstruction accuracy is assessed by LPIPS~\cite{zhang2018unreasonable} and SSIM~\cite{wang2004image}. 
In unpaired settings,  only FID~\cite{parmar2022aliased} and KID~\cite{binkowski2018demystifying} are reported due to the absence of ground truth. The high-level image similarity CLIP \cite{radford2021learning} is used in \cite{chen2024virtualmodelgeneratingobjectidretentivehumanobject}. To our knowledge, we are the first to attempt to build a comprehensive benchmark and enrich more novel and reliable metrics.

%-------------------------------------------------------------------------
\section{VTBench: Virtual Try-on Benchmark}
In this section, we detail the introduction of VTBench's components. Section 3.1 outlines the rationale behind designing the six evaluation dimensions, including their definitions and assessment evaluators. Section 3.2 elaborates on the test dataset collection process tailored for multidimensional analysis. Finally, in Section 3.3, we validate VTBench's alignment with human perception through dimension-specific preference annotations.

\subsection{Hierarchical Evaluation Dimensions}
We strongly claim that the motivation of the VTbench suite stems from the careful consideration of virtual try-on (VTON) in real-world scenarios as follows:

\noindent \textbf{1.) the deficiency of current  limited VTON evaluators}, existing evaluation metrics only FID and KID just calculates the distribution distance between two sets and severely affected by the number of data. They fail to give a judgment on one single image and usually suffer from the misalignment with human evaluations. 

\noindent \textbf{2.) the absence of the multidimensional explanations}, 
current VTON metrics merely simplify the performance of VTON generation models into a single numerical score. This oversimplification carries the risk of failing to reveal critical insights, such as identifying the specific strengths and weaknesses of individual models across different evaluation dimensions. 
This operation hinders the extraction of actionable insights for optimizing future architectural configurations and training paradigms from oversimplified unidimensional metrics. 
Consequently, moving beyond conventional single-numeric metric assessments of VTON generation quality, we propose a multidimensional evaluation framework that deconstructs the holistic concept of ``virtual try-on performance" into discrete, well-defined dimensions for granular analysis.

The ``virtual try-on quality" can be broken into six disentangled dimensions in a top-down manner with each evaluation dimension assessing one aspect of VTON generation quality. Considering the essence of virtual try-on tasks, it aims to modify the corresponding garment zones and remain other elements same with the model image. Thus, on the top level, we evaluate VTON performance from three broad perspectives: 1). General Image Quality, which focuses on the global perceptual quality and distribution similarity of the synthesized images. 2). Garment preservation, which delves into the logically reasonable substitution of the target garments, including the size and texture preservation. 3). Auxiliary consistency, which investigates the unchanged elements, such as the complex background and occluded hand, which are easily affected by the try-on models.

\subsubsection{General Image Quality}
We split ``General Image Quality'' into two finer aspects, ``Similarity'' and ``Aesthetics'', where the former only considers the general distribution similarity between the ground-truths and synthetic images, and the latter assesses  aesthetics and harmonization of global images. 

\noindent \textbf{Similarity}: We adopt the existing FID and KID to calculate the distribution distance of the deep feature space between model ground-truths and synthetic virtual try-on images. 

\noindent \textbf{Aesthetics}: 
The aesthetic score quantifies subjective experiences in virtual try-on results, including overall harmony (\textit{e.g.,}  integration between garments and human poses/backgrounds), color coordination, and style consistency, directly reflecting users' perception of "beauty." For instance, a virtually ill-fitting garment may be physically accurate yet receive a low
aesthetic score due to unnatural wrinkles or mismatched
lighting. Moreover, the aesthetic score captures higher-level
visual flaws, such as the presence of dissonance and 
anatomical distortions (\textit{e.g.,} unnatural body artifacts). To this end, we introduce aesthetic scores and calculate them based on the CLIP+MLP Aesthetic Score Predictor.

\subsubsection{Garment Preservation}
Despite advancements of recent emergence fancy paradigms, virtual try-on still faces significant garment-fitting challenges, particularly in two areas: (1) \textbf{Rich texture-aware maintenance}, where the transformation of intricate textures (\textit{e.g.,} patterns, text, stripes, trademarks) to the target model is hindered by the limitations of current structures and training manners; and (2) \textbf{Shape-aware fitting}, where clothing information leakage occurs in cross-category or shape-mismatched try-on scenarios, leading to generated garments that cover the entire agnostic mask region.   
To systematically assess model performance and robustness in addressing these challenges, we decompose ``Garment Preservation'' into two key dimensions— Size Fitness and Texture Fidelity—and develop corresponding evaluation metrics that align with human perceptual judgments.

\noindent \textbf{Texture Fidelity}: 
We identify that fine-grained texture transfer and consistent clothing appearance are crucial aspects of virtual try-on, yet there is a lack of suitable evaluation metrics in this domain. 
Existing metrics like SSIM \cite{wang2004image} have significant limitations. For example, even if the flowers on the target garment and the generated image are visually identical, a slight positional deviation can lead to a low metric score. Thus, we propose a font texture similarity (FTS) metric to evaluate the capability of font texture maintenance. Specifically, we introduce the OCR model \cite{li2022pp} for text feature extraction, and then use these features to customize a font-texture similarity metric to represent the capability of texture maintenance. Specifically, the text regions in both the clothing image  and the try-on result image can be obtained. 
Then we calculate the text similarity for each corresponding region via optimizing language processing distance criteria consisting of term frequency similarity and edit distance-based similarity components. 
Specifically, the component of edit distance-based similarity is:
\begin{equation}
\vspace{-1mm}
E_{_{edit}}(\boldsymbol{s}_{gt},\boldsymbol{s}_{p}) = 1-\frac{\text{LEV}(\boldsymbol{s}_{gt},\boldsymbol{s}_{p})}{max(|\boldsymbol{s_{gt}}|,|\boldsymbol{s_{p}}|)}
% \text{HA}(\mathcal{F}).
% \vspace{-1mm}
\end{equation}
% \vspace{-1mm}

where ${s}_{gt}$ is the text output of PPOCRv3 model from the target garment, and ${s}_{p}$ is the text output of PPOCRv3 model from the synthesis try-on image. $\text{LEV}$ denotes the Levenshtein distance from the string  ${s}_{p}$ to the string ${s}_{gt}$.The part of term frequency similarity is shown as: 
% \vspace{-1mm}
\begin{equation}
% \vspace{-1mm}
E_{_{term}}(\boldsymbol{s}_{gt},\boldsymbol{s}_{p}) = \frac{\text{TF}(\boldsymbol{s}_{gt}) \cdot \text{TF}(\boldsymbol{s}_{p})}{|| \text{TF}(\boldsymbol{s_{gt}})||\times||\text{TF}(\boldsymbol{s_{p}}||)}
% \text{HA}(\mathcal{F}).
% \vspace{-1mm}
\end{equation}
% \vspace{-1mm}
where $\text{TF}$ is the term frequency calculation, and $E_{_{term}}$ calculates the Cosine Similarity of the term frequency between the string ${s}_{gt}$ and ${s}_{p}$.
The true positive \text{TP}, false positive \text{FP}, and false negative \text{FN} predictions of all font characters are listed as \text{TP}$_s=|s_{gt}|\cap|{s}_{p}|$, \text{FP}$_s=|{s}_{p}|-|s_{gt}|$, and \text{FN}$_s=|s_{gt}|-|{s}_{p}|$. $||$ means the recognized font character code. Precision $E_p$, Recall $E_r$, and F1-Measure $E_f$ are calculated to analyze the accuracy of font character recognition. 

Our proposed text semantic-texture similarity is finally formulated as:
\begin{equation}
\begin{aligned}
% \vspace{-1mm}
E_{TSS}=& w_{1}*\frac{1}{N} \sum\limits_{i=1}^{N} E_{term}^{i}+w_{2}*\frac{1}{N} \sum\limits_{i=1}^{N}E_{edit}^{i} \\ & +w_{3}*\frac{1}{N} \sum\limits_{i=1}^{N}(E_{p}^{i}+E_{r}^{i}+E_{f}^{i})
\end{aligned}
\end{equation}
where the weights of $w_{1}$, $w_{2}$ and $w_{3}$ are set to 0.2. $i$ and $N$ are the \textit{i}-th region and the total number of regions, respectively. The average similarity score across multiple regions is considered as the final similarity score.

\noindent \textbf{Size Fitness}: Virtual try-on methods only handle simple scenarios (\textit{e.g.,} long sleeves $\leftrightarrow$ short sleeves) and struggle with complex cases such as long skirts $\leftrightarrow$ upper jackets. The difficulty of cross-category try-on lies in the lack of logical reasoning about how the target garment is dressed on the model, specifically the screen into the different zones, \textit{i.e.,} the reconstruction zone that is learning from the model image, the try-on zone that is referring to the garment image, and imagination zone that developing the imaginative potential of Diffusion to paint on. To quantify this cross-category try-on ability strongly needed by real users and personalized styling demands, we introduce Vision Language Models (VLM) to analyze the logical reasoning about how the target garment is dressed on the model. Given a model image, target garment, and the virtual try-on result, we introduce Qwen-VL-Max to precisely judge whether the garment try-on result is shape-fitting obeying the cross-category logical reasoning. Then, we utilize this VLM model to calculate the ratio of the number of correct results to the total number of samples in the entire test set as the accuracy of current try-on model in cross-category try-on.  
\begin{equation}
E_{_{size}}=\frac{S_{g}[\text{VLM}(\text{Concat}(I_m,I_g, I_{syn}))]}{S_{\text{cases}}} 
\end{equation}
where $S_{g}$ is the number of reasonable cross-category cases, $S_{cases}$ is the total cases. The model image $I_m$, garment image $I_g$, and synthetic try-on image $I_{syn}$ are concatenated as the inputs to the Qwen-VL-Max \cite{bai2025qwen2} model.

\subsubsection{Auxiliary Consistency}
The fundamental objective of virtual try-on methodologies lies in accurately identifying modifiable regions while preserving invariant areas within the model image. The aforementioned garment evaluation metrics specifically assess the visual quality of synthesized garments in the target region. This section systematically evaluates two critical aspects: preservation consistency of unmodified regions, and contextual coherence in complex scenarios. Real-world virtual try-on applications typically involve cluttered backgrounds with frequent human-object interactions, presenting significant challenges for background preservation. Furthermore, hand-occluded garment scenarios, which frequently occur in real-world applications involving complex human poses, introduce additional complications for garment synthesis algorithms. To comprehensively evaluate model performance and robustness in addressing consistency challenges, we mainly disentangle ``Auxiliary Consistency'' into two aspects including Background Consistency and Hand Consistency. For each dimension, we develop quantitative evaluation metrics that are both computationally rigorous and perceptually aligned with human visual assessment.

\noindent \textbf{Background Consistency}:
In virtual try-on for real-world scenarios, the background is often highly complex, involving interactions between people and objects. However, the current inpainting-based virtual try-on paradigm tends to alter background information to some extent. As an ideal garment replacement algorithm, it should minimize modifications to reasonable areas while preserving background consistency as much as possible. Additionally, any altered background regions should maintain harmony with the surrounding context. Then we respectively calculate the pixel-wise  and semantic-wise similarity after determining the target mask zone.
$M_{GT}$  and $M_{syn}$ are the human mask of model images and synthetic try-on results, respectively. We dilate the maximum zone and obtain the residual mask zone $M_{_{D}}$ around human as follows: 
\begin{equation}
M_{_{D}}=\text{Dilate}[(M_{GT},M_{syn})_{\text{max}}]-M_{GT}
\end{equation}
Afterwards, the pixel-level of background consistency is calculated based on L1 distance around $M_D$ zone as follows:
\begin{equation}
E_{pixel}={|I_{syn}-I_{m}|}_{\text{L1}, M_{D}}
\end{equation}
Besides, the semantic-wise similarity of background consistency is obtained based on DINO model around $M_D$ residual area.
\begin{equation}
E_{semantic}={|I_{syn}-I_{m}|}_{\text{DINO}, M_{D}}
\end{equation}

\noindent \textbf{Hand Consistency}:
Current virtual try-on systems typically require models to maintain rigid poses with hands positioned strictly at their sides, ensuring garments remain completely unobstructed. This artificial constraint significantly limits the practical applicability of existing methods in real-world scenarios. The challenge is particularly severe in digital human applications, where natural poses frequently involve complex hand gesture language that occluded garment. 
Such occlusions contribute a lot of failure cases for contemporary virtual try-on algorithms. It requires high-accommodation of different pose variations while maintain the hand appearance and structure consistent especially on hand-occluded cases. We introduce the joint position error (JPE) to get the each joint distance between the synthetic and model images as the final error:
\begin{equation}
E_{\text{MPJPE}}=\frac{1}{N_s} \sum\limits_{i=1}^{N_{s}} {\Vert m_{syn, s}(i)- m_{model, s}(i)\Vert }_{2}
\end{equation}
where $m_{syn,s}(i)$  and $m_{model,s}(i)$ are the functions that return the coordinates of the i-th joint of skeleton from the synthetic try-on images and the model images, respectively.  $N_s$ is the number of the joints in skeleton S.

\subsection{Test Dataset Collection and Analysis}
As an evaluator suite of each hierarchical dimension, it not only requires the custom-designed metrics but only cleaning the whole dataset to focus on the specific dimension which requiring greatly intensive labor annotations. Specifically, we collect an additional dataset of 50,000 images sourced from online retail sites. Considering just varying the specific dimension without involving other dimension disturbance, we carefully filter the whole dataset and classify the remaining dateset into the four classes catering for each well-disengaged evaluation dimension. 

\noindent \textbf{Complex Background Consistency Dataset (CBC)}: 
The fundamental significance of complex background testing lies in transitioning garment virtual try-on algorithms from `controlled lab environments' to `real-world deployment scenarios.' Only by passing rigorous validation in complex settings can virtual try-on technology genuinely serve daily life and evolve into a reliable `digital fitting room' for consuming users. We filter these data into the complex scenarios including the street scene, wild scene, and none-simple indoor scene, and collect 5,00 images for further prepossessing and evaluation.

\noindent \textbf{Font Texture Fidelity Dataset (FTF)}:
Font Texture Fidelity Dataset focuses on evaluating the model's ability to accurately transfer fine-grained textures of the font.  In the FTF set, we observe that text structures serve as reliable indicators of the model's fine-grained texture transfer capabilities. We manually annotate and collect 600 images for further prepossessing and evaluations.

\noindent \textbf{Cross-category Size-fitting (CSF) Dataset}:
To comprehensively evaluate cross-category virtual  try-on performance, we construct the Cross-category Size-fitting Dataset (CSF) by systematically combining models and garments from four distinct categories in DressCode: top, lower, dresses, skirts. The dataset contains 400 base model-garment pairs with 100 pair category), which are strategically combined to create 1,000 test pairs that rigorously assess cross-category compatibility.

\noindent \textbf{Hand-occluded Consistency Dataset (HOC)}: 
Current virtual try-on datasets predominantly feature hands separated from the body, resulting in a scarcity of data for arbitrary hand poses and challenging hand-occlusion scenarios. To address this critical gap in real-world applications, we introduce a novel dataset focusing on complex hand poses that occlude garments. Our pipeline begins by processing images using a parsing model alongside the HaMeR model \cite{pavlakos2024reconstructing} to detect hand bounding boxes and identify occlusion instances. Rigorous manual inspection was conducted to eliminate erroneous or low-quality entries, ensuring the dataset’s reliability.
From an initial pool of 50,000 images collected from online retail platforms, we curated the Hand-occluded Consistency Dataset (HOC), a specialized test set comprising 1,433 high-quality images. This dataset enables targeted evaluation of virtual try-on methods, specifically their robustness in handling occlusions.

\subsection{Human Preference Annotation}
We perform human preference labeling on massive try-on synthetic results, which verifying the evaluation's alignment with human perception in each of the six evaluation dimensions. 
Given a garment $g_{i}$, and six image virtual try-on models [{$M{1},M{2},M{3},M{4},M{5},M{6}$}], these VTON models are regarded as a group. This group yields five-teen binary combinations, and ask human annotators to indicate their preference (\textit{i.e.,} A or B is better or equal). Each dimension takes 20 samples, and total it provides $N$$\times$5$\times$15 image comparisons. Human annotators follow the rules that only consider the specific evaluation dimension without disturbances of other irrelevant dimension.

\begin{table*}[t!]
  \centering
  % \footnotesize
    \small 
    % \scriptsize
    % \tiny
  \caption{\small 
   VTBench quantitative  results of 15 SOTA methods  on different network paradigm on our benchmark datasets of each dimension. The best results of all baselines
   are highlighted in \underline{\textbf{underline}}, and the best results of each specific framework (\textit{i.e.,} GAN, UNet-based or DiT-based Diffusion model) is marked in \textbf{bold}. 
  }\label{tab:main_tab}
  % \vspace{-10pt}
  \renewcommand{\arraystretch}{1} 
  \resizebox{1\textwidth}{!}{
  \begin{tabular}{l||ccc|cc|cc}
  %\hline
  \toprule
    & \multicolumn{3}{c|}{\tabincell{c}{Image Virtual Try-on Quality }} & \multicolumn{2}{c|}{\tabincell{c}{ Garment Preservation
}} & \multicolumn{2}{c}{ \tabincell{c}{Auxiliary Consistency}}  \\
   \cline{2-8}
    ~~~ &  \multicolumn{2}{c}{ \tabincell{c}{Similarity}}    & Aesthetics  & Texture  &Size  & Background & Hand      \\ 

    Baseline Models~~~ &  FID $\downarrow$ & KID $\downarrow$ & $A_{s}$ $\uparrow$ & $F_{t}$ $\uparrow$  & $\text{VLM}_{s}$ $\uparrow$ & $B_{consistency}$ $\downarrow$ & $H_{consistency}$ $\downarrow$     \\
    \midrule
    \multicolumn{8}{c}{ \textit{GAN-based} }\\
     \midrule
    2021 PF-AFN ~\cite{ge2021parser} & 74.7482&	40.8694&3.7881 & 0.3278& - & 4109.3963& 75.4443 \\
    2022 FS-VTON ~\cite{he2022style} & 71.0491 &40.7414	&3.9055 &0.3165 & - & 2940.1240& 42.8215 \\
    2022 DA-FLOW ~\cite{bai2022single} & 100.2046&64.2756&3.8198 &0.2690 & - & \textbf{2686.7483}&179.8285\\
    2022 HR-VITON ~\cite{lee2022high} & 56.5964&23.1358&4.5128 & 0.4099& - &6495.7036 & 43.4961 \\
    2024 SD-VITON ~\cite{shim2024towards} &\textbf{49.6304}&	\textbf{16.5367} &\textbf{4.5716} & \textbf{0.4786} & -  & 6070.9095&\textbf{38.6793}\\
        
    \midrule
    \multicolumn{8}{c}{ \textit{UNet-based Diffusion} } \\
     \midrule
    2023 LaDI-VTON ~\cite{morelli2023ladi} &31.4483	& 7.1166 & 5.0518  &0.0601 & - & 485.8788 & 45.7704\\
    2024 StableVTON   ~\cite{kim2024stableviton} &\ \underline{\textbf{27.8368}}&	\underline{\textbf{2.2319}}	&4.9862 & 0.2009 & - & \underline{\textbf{132.7260}} & 40.2104 \\
    2024 TPD ~\cite{yang2024texture} & 32.6199&	6.7273&4.8379 & 0.3383& - &161.3437 &39.4060\\
    2024 CAT-DM ~\cite{zeng2024cat} & {28.7471}&	4.4718&	4.9188 & 0.0140& 12.8571 & 154.1549&42.4447\\
    2024 IDM-VTON  ~\cite{choi2024improving} &29.8710 &4.9117	&\underline{\textbf{5.1243}} & 0.3387 & 15.7143 & 173.7212 &48.3768\\
    2024 OOTD ~\cite{xu2024ootdiffusion} & 45.9138	&16.7691&	5.0367 & 0.2153 & 13.5714 & 205.4843 &62.5795\\
    2024 CatVTON ~\cite{chong2024catvton} & 29.2677 &	3.4891	&5.0336 & \textbf{0.5433} & 21.4286 &176.5129 &35.7663\\
    2024 VTON-HandFit ~\cite{liang2024vton} & 28.9344	&2.4419	&5.1170  & 0.4949 & \textbf{29.0476}& 319.8210 & \underline{\textbf{15.4596}}\\
    \midrule
    \multicolumn{8}{c}{ \textit{DiT-based Diffusion}} \\
     \midrule
    2024 FitDit ~\cite{jiang2024fitdit} & \textbf{29.6194}&	\textbf{3.6564}&	\textbf{5.1135}  & \underline{\textbf{0.6481}} & \underline{\textbf{55.2381}}&\textbf{137.4822} &33.7351\\
    2025 CrossVTON ~\cite{luo2025crossvton} & 32.0386&3.9203 &5.1018 & 0.5728 & 48.0952 & 715.6857& \textbf{25.2247} \\
  
  \toprule
  \end{tabular}
 }
  \vspace{-5mm}
\end{table*}

\section{Experiments}
We introduce the most-recent 15 baselines including GAN-based (PF-AFN \cite{ge2021parser},  FS-VTON \cite{he2022style}, DA-FLOW~\cite{bai2022single},  HR-VITON ~\cite{lee2022high},  SD-VITON ~\cite{shim2024towards}), U-Net-based Diffusion (LaDI-VTON~\cite{morelli2023ladi}, StableVTON~\cite{kim2024stableviton}, TPD~\cite{yang2024texture}, CAT-DM~\cite{zeng2024cat}, IDM-VTON~\cite{choi2024improving}, OOTD~\cite{xu2024ootdiffusion}, CatVTON~\cite{chong2024catvton}, VTON-HandFit~\cite{liang2024vton}), and DiT-based Diffusion models (FitDit~\cite{jiang2024fitdit}, CrossVTON~\cite{luo2025crossvton}) for VTBench evaluations. For a fair comparison, all results are reproduced by official open-source models with the recommended setting.
% \subsection{Settings}

\begin{wrapfigure}{r}{0.48\textwidth}
\includegraphics[width=1\linewidth]{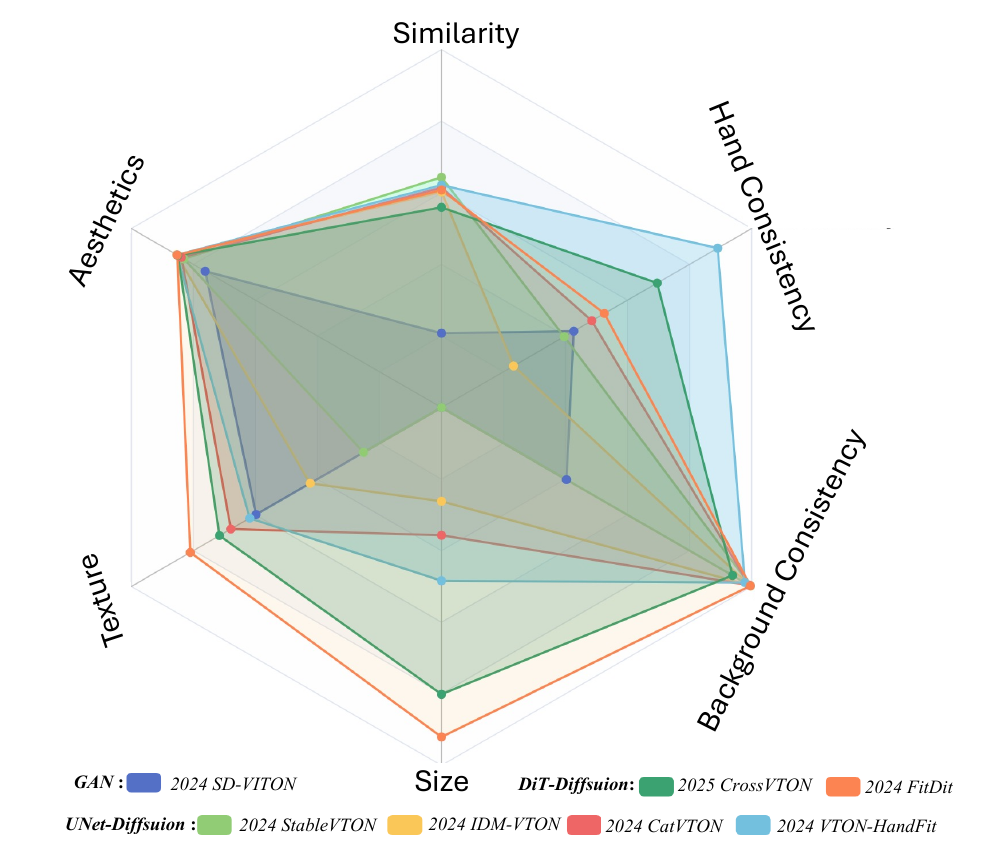}
\vspace{-5mm}
\caption{\footnotesize{VTBench Evaluation Results of SOTA Virtual try-on Models including GAN, UNet-based and DiT-based Diffusion.}}
\label{fig:fig_rad}
\vspace{-12mm}
\end{wrapfigure} 

\subsection{Quantitative Results}
In Tab. \ref{tab:main_tab} and Fig. \ref{fig:fig_rad}, we classify all baselines based on network structures into three classes (\textit{i.e.,} GAN-based, UNet-based, and DiT-based Diffusion). 
The UNet-based Diffusion (StableVTON) obtains the similar similarity, aesthetics, and background consistency score than DiT-based structure (FitDit), but significantly better than GAN-based structure (SD-VITON) baselines. Such an observation indicates that the Diffusion-based models perform its greater superiority than GAN-based methods on the high-fidelity image generation capability bridging the gap between the synthetic and real image. 
FitDit, a DiT-architected diffusion model, outperforms GAN and UNet-based diffusion frameworks by notable margins in maintaining garment structural integrity, particularly in texture detail conservation and size fitting.

% DiT-based ()and UNet-based structures. 
% \noindent \textbf{Performance of DiT-based Diffusion models}
% \noindent \textbf{Performance of UNet-based Diffusion models}
% \noindent \textbf{Performance of GAN-based models}
% \subsection{Qualitative Results}
% \subsection{Metric Hypermeter Analysis}
% \subsection{Dataset Analysis and Demonstration}
% \subsection{Validating Human Alignment of VTBench}

\subsection{Validating Human Alignment}
\begin{figure}[htbp]
\centering
\includegraphics[width=1\textwidth]{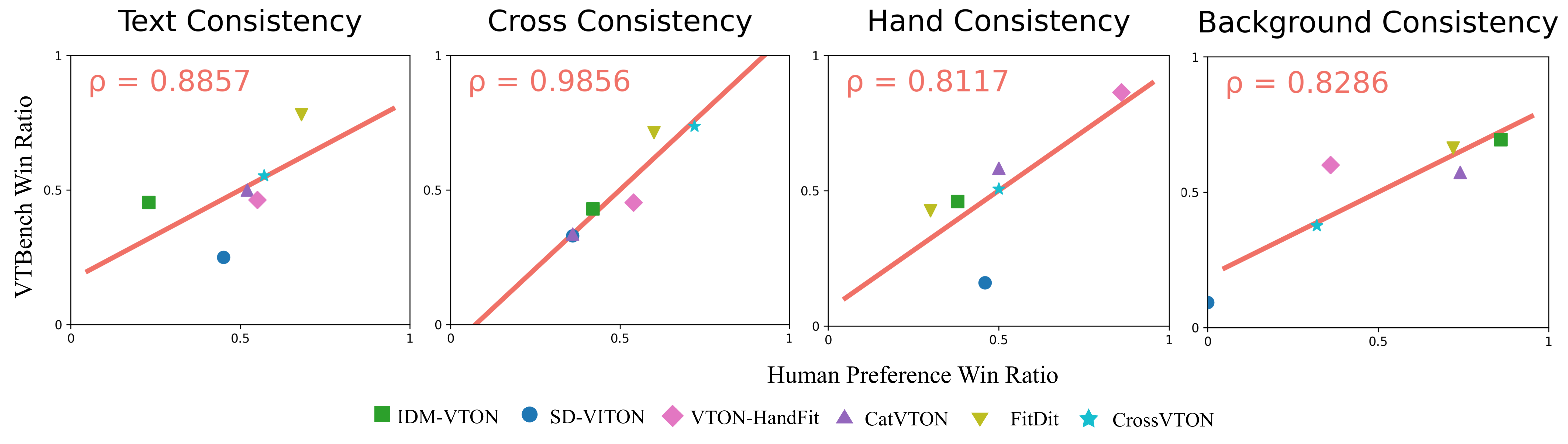}
\vspace{-7mm}
\caption{\textbf{Validate VTBench’s Human Alignment}.
Our experimental results demonstrate that VTBench evaluations across all dimensions exhibit a strong alignment with human perceptual judgments. Each plot illustrates the verification results for a specific VTBench dimension, where a single dot represents the human preference win rate (x-axis) and the VTBench evaluation win rate (y-axis) for a given virtual try-on generation model. To assess the correlation, we perform a linear regression analysis and compute the Spearman’s rank correlation coefficient for each dimension.
}
\label{fig:human_eva}
% \vspace{-3mm}
\end{figure}

To verify that our evaluation method accurately captures human perceptual judgments, we conducted a large-scale human annotation study for each dimension. The correlation between VTBench evaluation results and human preference annotations is illustrated in Fig. \ref{fig:human_eva}.
\noindent \textbf{Win Ratio:} 
Following the VBench video quality benchmark framework, we compute each model's win ratio based on human annotations. In pairwise comparisons: 1). A model receives 1 point if its output is preferred over the other's, while the competing model receives 0. 2). In case of a tie, both models are awarded 0.5 points. The win ratio for each model is then derived by dividing its total score by the number of pairwise comparisons it participated in. 

\noindent \textbf{Per-Dimension Evaluation Analysis:}
For each evaluation dimension, we compute model win ratios using both (1) VTBench automated evaluations and (2) human annotation results, then analyze their correlation in Fig. \ref{fig:human_eva}. Our findings demonstrate \textbf{strong alignment} between VBench's per-dimension assessments and human preference judgments.

\section{Insights and Discussions}
\noindent \textbf{\textit{DiT outperforms others on garment preservation:}} To increase the receptive field of network and get global knowledge, UNet adopts the downsampling and upsampling operation to focus more attention on global but not high-resolution latent features.  
For SD v1.5, the attention-related parameter ratios at latent resolutions higher than 64 $\times$ 48 only account 16\% \cite{jiang2024fitdit}. This focus undermines the maintenance of high-resolution, rich textures in garments, which is crucial for achieving high-fidelity virtual try-on. In contrast, SD3 discards the downsampling and upsampling operations, and shifts more attention on the high-resolution latent features (\textit{i.e.,} allocates over 99\% of its parameters to the 64 $\times$ 48 resolution). Therefore, for tasks such as virtual try-on, which demand high fidelity in detail preservation, the DiT architecture emerges as a superior choice.

\noindent \textbf{\textit{Similarity dimension fails to represent texture maintenance and size-fitting:}} As the comprehensive metrics of unpaired image virtual try-on similarity quality, FID and KID fail to reflect the garment texture preservation ability, which stands for the vital texture preservation dimension of virtual try-on. As shown in Tab. \ref{tab:main_tab}, StableVTON earns the best FID and KID similarity score but gets a much lower texture preservation score than FitDit, which indicates that the FID and KID similarity only demonstrates the distribution similarity between two batch images but is not suitable acting as the texture consistency indicator. Besides, VTON-HandFit gets much better FID and KID scores but also shows its overwhelming inferiority in garment size preservation than the FitDit model. 
Therefore, it is necessary to propose our garment texture and size preservation metric to correctly evaluate the garment preservation of virtual try-on tasks.

\noindent \textbf{\textit{More attention on garment preservation metrics:}} As a virtual try-on task, it is the absence of reasonable metrics as unpaired metrics to evaluate the garment preservation, mainly consisting of texture and size.  After collecting the paired data, previous works usually adopt the manner that directly replace the model wearing the target garment with the target garment itself. The availability of difficulty of paired try-on data limit its in-time evaluation in out-of-domain data of complex real-world scenarios. Above human alignment analysis verifies the effectiveness of proposed garment preservation metrics, and we call on that more attention should be assigned to these garment preservation metrics to judge the garment-relevant texture and size quality.

\noindent \textbf{\textit{GANs exhibit limitations in maintaining consistent background:}}  As shown in Tab. \ref{tab:main_tab}, GAN-based methods perform much worse than diffusion-based methods in the pixel-wise background consistency. Basically, Diffusion-based studies have treated virtual try-on as an exemplar-based image inpainting problem. Besides, GAN-based also adopts the try-on conditions generator and then fuses these conditions (\textit{e.g.,} clothing-agnostic, warped clothing image, pose map) for try-on image generation as the inpainting task. 
We suspect that it mainly comes from the foundation model discrepancy that Diffusion-based studies perform much stronger and harmonious inpainting capability than GAN-based models. Then it leads to the intersection zone between individual and background more closer to the original one. In addition, the performance of background consistently using DiT-based Stable Diffusion-3 model is similar to the results based on the UNet-based stable diffusion models, which indicates that the upgrade of UNet-to-DiT Diffusion structure does not dramatically boost better background consistency.

\noindent \textbf{\textit{Hand-priors benefit the hand-occluded reconstruction:}}
As shown in Tab. \ref{tab:main_tab}, VTON-HandFit achieves the best score and dramatically lowers the hand-consistency error than the second-best FitDit by 38.7\%.
VTON-HandFit adopts the encoding of structure-parametric and visual-appearance priors to solve hand pose occlusion problems.
From Tab. \ref{tab:main_tab}, such hand prior embedding is verified to be effective in reconstructing the occluded hand in virtual try-on tasks. We recall that this hand-prior embedding probably facilitates the virtual try-on results, especially on hand-occluded scenarios.
\vspace{-2mm}
\section{Conclusion}
\vspace{-2mm}

As the virtual try-on generation garners increasing attention, a systematic and rigorous evaluation framework is crucial to assess the progress of existing models and steer future research directions. To address this need, we propose VTBench, the first comprehensive benchmark suite designed specifically for evaluating virtual try-on generation models.
VTBench is characterized by its multi-dimensional evaluation criteria, human-aligned assessment methodology, and insight-rich analysis, making it a powerful tool for benchmarking current and future advancements in the field. 

\noindent \textbf{Broader impacts and Limitations:} 
VTBench will serve as a foundational resource for researchers, facilitating more objective comparisons and inspiring further innovation in virtual try-on generation. We believe this work represents a significant contribution to both the virtual try-on generation community and the broader field of AI-driven fashion technology.
But VTBench currently focuses exclusively on the most prevalent image virtual try-on task. 
It does not support video-based virtual try-on evaluations, particularly in assessing temporal smoothness.
{
    \small
    \bibliographystyle{unsrt}
    \bibliography{neurips_2025}
}

\end{document}